\documentclass[conference]{IEEEtran}

\usepackage{amsmath,amssymb}
\usepackage{graphicx}
\usepackage{booktabs}
\usepackage{multirow}
\usepackage{hyperref}
\usepackage{cite}
\usepackage{enumitem}
\usepackage[T1]{fontenc}

\begin{document}

\title{Shower-Aware Dual-Stream Voxel Networks for\\Structural Defect Detection in Cosmic-Ray Muon Tomography}

\author{
\IEEEauthorblockN{Parthiv Dasgupta, Sambhav Agarwal, Palash Dutta, Raja Karmakar, Sudeshna Goswami}
\IEEEauthorblockA{Department of Computer Science and Engineering\\
Heritage Institute of Technology, Kolkata, India\\
Email: dasguptaparthiv@gmail.com}
}

\maketitle

\begin{abstract}
Cosmic-ray muon tomography enables non-destructive inspection of reinforced concrete, yet conventional reconstruction algorithms such as POCA cannot reliably separate structural defects from the scattering signature of steel reinforcement. We present a Shower-Aware Dual-Stream Voxel Network (SA-DSVN) that jointly processes muon scattering kinematics (9 channels) and secondary electromagnetic shower multiplicities (40 channels) to perform six-class voxel-level segmentation on a $20^3$ grid. Training data were generated with \textit{Vega}, a cloud-native Geant4 simulation framework that produced 4.5 million muon events across 900 volumes containing four clinically relevant defect types---honeycombing, shear fracture, corrosion voids, and delamination---embedded within a dense $7 \times 7$ rebar cage. A systematic ablation study over five architectural variants reveals that the shower multiplicity stream alone accounts for the majority of discriminative power, raising defect-mean Dice from 0.535 (scattering only) to 0.685 (shower only). On 60 independently simulated validation volumes unseen during training, the model achieves 96.3\% voxel accuracy, per-defect Dice scores of 0.59--0.81, and 100\% volume-level detection sensitivity across all four defect classes at an inference cost of 10\,ms per volume. These results establish secondary shower multiplicity as an effective feature for learned muon tomographic reconstruction and demonstrate that physics-informed data augmentation is essential for generalisation beyond the training distribution.
\end{abstract}


\section{Introduction}
\label{sec:intro}

Reinforced concrete degrades. Rebar corrodes, voids form around aggregate, shear cracks propagate under cyclic loading. Left undetected, these defects compromise structural integrity long before visible damage appears on the surface. Standard non-destructive testing---ground-penetrating radar, ultrasonic pulse velocity, X-ray radiography---either lacks the penetration depth to image through heavily reinforced sections or requires controlled radiation sources that limit field deployment.

Cosmic-ray muon tomography sidesteps both limitations. Natural atmospheric muons penetrate metres of concrete and steel; their Coulomb scattering angles encode the density distribution of the traversed material. No artificial source is needed. The detector hardware is passive and safe for continuous field operation. Yet after two decades of development~\cite{alvarez1970,borozdin2003}, muon tomography remains largely a laboratory technique. The reason is algorithmic, not physical.

The dominant reconstruction method, Point of Closest Approach (POCA), estimates scattering locations by intersecting incoming and outgoing muon trajectories. In simple geometries---a tungsten cube inside an air box---POCA works well. In real infrastructure, the picture breaks down. A typical bridge pier contains a dense cage of steel rebar spaced at 100--200\,mm intervals. Every bar produces strong scattering that POCA cannot distinguish from a nearby void or crack. The result is a noisy point cloud in which genuine defects are buried under structural false positives. Statistical methods such as Maximum Likelihood Scattering and Density (MLSD)~\cite{schultz2007} improve contrast but require hours of iterative computation per volume and still struggle with the rebar-defect ambiguity.

A separate physical signal has been overlooked. When a muon traverses high-$Z$ material, it produces secondary electromagnetic showers---knock-on electrons, bremsstrahlung photons---at rates that scale with atomic number. Steel ($Z=26$) generates measurably more secondaries than concrete ($Z_{\text{eff}} \approx 11$) or air. This multiplicity difference is recorded by the same detector planes that measure scattering angles, yet no existing reconstruction algorithm exploits it.

We exploit it. This paper introduces the Shower-Aware Dual-Stream Voxel Network (SA-DSVN), a 3D convolutional architecture that processes scattering kinematics and shower multiplicities as two independent input streams before fusing them via cross-attention. Trained on synthetic data generated by \textit{Vega}, our cloud-native Geant4~\cite{agostinelli2003,allison2016} simulation framework, the network performs six-class voxel segmentation---concrete, rebar, honeycombing, shear fracture, corrosion void, and delamination---on a $20^3$ volumetric grid.

The contributions of this work are:

\begin{enumerate}[leftmargin=*]
    \item \textbf{A dual-stream architecture} that separates scattering and shower features, enabling the network to learn the rebar signature independently from density anomalies. Ablation experiments show the shower stream alone raises defect-mean Dice by 28\% over scattering alone (0.685 vs.\ 0.535).

    \item \textbf{A validated simulation-to-inference pipeline.} \textit{Vega} generated 4.5 million muon events across 900 reinforced concrete volumes with four embedded defect types. An additional 60 volumes from an independent simulation campaign confirm that the trained model generalises, achieving 96.3\% voxel accuracy on data it has never seen.

    \item \textbf{Evidence that data augmentation is critical for generalisation} in voxelised muon tomography. Without augmentation, defect segmentation collapses on held-out data (Dice $\to$ 0). With 3D spatial flips and intensity perturbations, the same architecture recovers to Dice 0.59--0.81 across defect classes.
\end{enumerate}

The remainder of this paper is organised as follows. Section~\ref{sec:related} surveys prior work in muon reconstruction and learned segmentation. Section~\ref{sec:vega} describes the Vega simulation framework and the data generation pipeline. Section~\ref{sec:method} details the SA-DSVN architecture, loss function, and training protocol. Section~\ref{sec:experiments} presents the ablation study, generalisation tests, and augmentation analysis. Section~\ref{sec:conclusion} concludes with limitations and future directions.

\section{Related Work}
\label{sec:related}

\subsection{Scattering-Based Muon Reconstruction}

The foundation of muon tomography was laid by Borozdin et al.~\cite{borozdin2003}, who demonstrated that cosmic-ray muons could image high-$Z$ objects concealed within cargo containers. Their approach relied on measuring the angular deflection of muons passing through a target and attributing each scatter to a single spatial point via POCA. Schultz et al.~\cite{schultz2007} replaced this geometric heuristic with a statistical framework, treating the reconstruction as a maximum-likelihood problem over a discretised voxel grid. Both methods assume that scattering angle is the sole observable. In the presence of dense steel reinforcement, this assumption breaks: rebar and defects produce overlapping scattering distributions, and the algorithms cannot separate them without prohibitively long exposure times.

\subsection{Machine Learning in Muon Tomography}

Attempts to apply learned models to muon data remain sparse. Tripathy et al.~\cite{tripathy2021} used statistical feature extraction from scattering distributions to classify simple geometric targets but did not attempt voxel-level reconstruction. Pezzotti et al.~\cite{pezzotti2025} trained a convolutional network to enhance 2D muon radiography images, improving contrast for cargo screening. Their work operates on projected images, not volumetric data, and targets a fundamentally different task (threat detection in shipping containers vs.\ structural defect identification in concrete). Huang et al.~\cite{huang2022} applied gradient boosting to classify material types from scattering features but again limited the scope to binary or ternary classification rather than spatial segmentation.

No prior work has attempted multi-class 3D voxel segmentation of structural defects within reinforced concrete using muon tomography data. Nor has any prior work exploited secondary shower multiplicities as a learned feature.

\subsection{3D Medical and Industrial Segmentation}

The encoder-decoder paradigm for volumetric segmentation is well established in medical imaging. Ronneberger et al.~\cite{ronneberger2015} introduced U-Net for 2D biomedical segmentation; Milletari et al.~\cite{milletari2016} extended it to 3D with V-Net and proposed the Dice loss to handle class imbalance. Attention mechanisms were incorporated by Oktay et al.~\cite{oktay2018} through attention gates that suppress irrelevant encoder features during decoding. Our architecture draws on these ideas but differs in two respects: the input is not a raw image but a physically derived multi-channel feature tensor, and the dual-stream design has no analogue in medical segmentation where a single imaging modality is typical.

\subsection{Synthetic Data and Domain Randomisation}

Training on simulation data is standard practice when real labelled volumes are unavailable. Tobin et al.~\cite{tobin2017} showed that randomising visual parameters during simulation (domain randomisation) transfers learned representations to real-world robotics tasks. We adopt a similar philosophy: Gaussian detector noise, alignment jitter, and stochastic defect placement are applied during data generation, while 3D spatial augmentation is applied during training. Our ablation on augmentation (Section~\ref{sec:aug}) quantifies the effect directly.

\section{Vega Simulation Framework}
\label{sec:vega}

Supervised voxel segmentation requires paired inputs (detector hits) and ground-truth labels (material identity at every voxel). No such dataset exists for reinforced concrete under muon irradiation. We built one.

\subsection{Physics Engine}

\begin{figure}[t]
    \centering
    \includegraphics[width=\columnwidth]{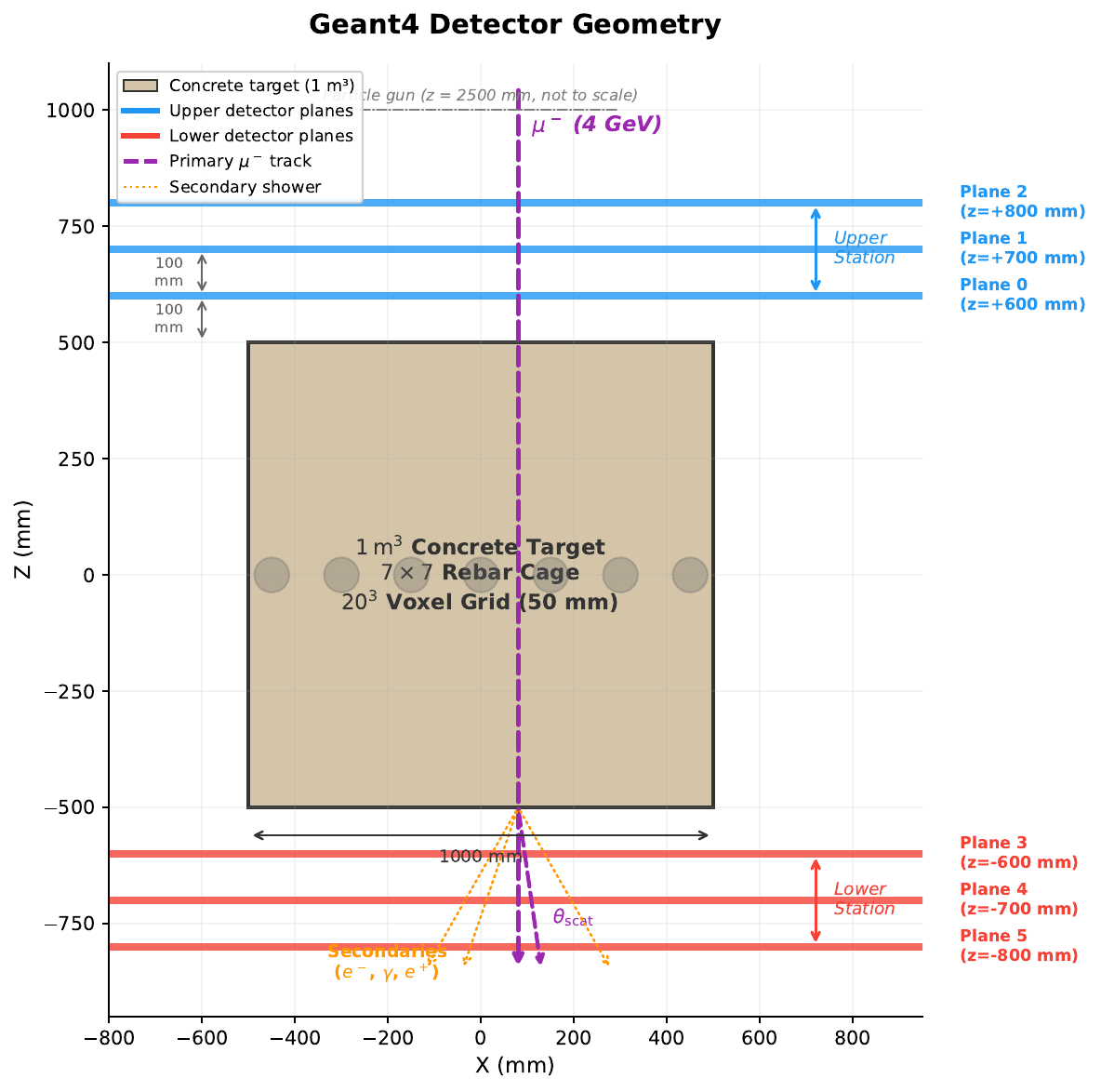}
    \caption{Side-view schematic of the Geant4 detector geometry. A 1\,m$^3$ concrete target containing a $7\times7$ rebar cage sits between two detector stations (3 planes each, spaced 100\,mm apart). A 4\,GeV $\mu^-$ enters from above; scattering angle and secondary shower particles are recorded on all six planes.}
    \label{fig:detector_geometry}
\end{figure}

Vega wraps Geant4 v11.x with the \texttt{QGSP\_BERT} physics list, which handles Coulomb scattering, ionisation, bremsstrahlung, pair production, and hadronic cascades. A planar particle gun at $z = 2500$\,mm fires mono-energetic muons at 4\,GeV with vertical incidence ($p_z = -1$). Each simulation job processes 5,000 muon events through the target geometry and records hits on six tracking planes---three above and three below the specimen---capturing position, momentum, energy deposit, particle species, and timing for every primary and secondary particle. The full detector geometry is illustrated in Fig.~\ref{fig:detector_geometry}.

\subsection{Target Geometry}

\begin{figure*}[t]
    \centering
    \includegraphics[width=\textwidth]{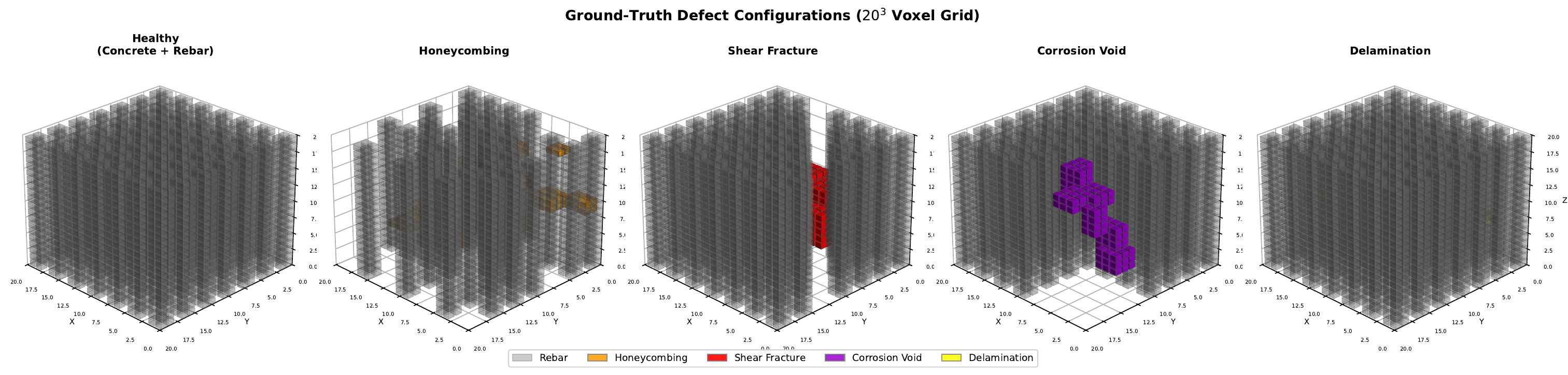}
    \caption{Ground-truth voxel label maps for the five target configurations on the $20^3$ grid. Grey voxels represent the $7\times7$ rebar cage (present in all configurations). Coloured voxels indicate defect regions: orange = honeycombing (scattered bar removal), red = shear fracture (diagonal section removal), purple = corrosion void ($3\times3$ corner removal), yellow = delamination (thin air layer). Concrete voxels are rendered transparent.}
    \label{fig:defect_cocktail}
\end{figure*}

The target is a $1\,\text{m}^3$ concrete block ($\rho = 2.3\,\text{g/cm}^3$) containing a 7$\times$7 grid of vertical steel rebar ($\rho = 7.87\,\text{g/cm}^3$, radius 15\,mm, spacing 150\,mm). This cage is present in every volume and serves as the primary source of background scattering. Four defect classes are embedded individually into the cage (Fig.~\ref{fig:defect_cocktail}):

\begin{itemize}[leftmargin=*]
    \item \textbf{Honeycombing:} stochastic removal of 40\% of the rebar grid, simulating distributed aggregate segregation and air voids.
    \item \textbf{Shear fracture:} removal of a diagonal section of the cage, representing stress-induced failure planes.
    \item \textbf{Corrosion void:} removal of a localised $3 \times 3$ corner section of the rebar grid, modelling water-ingress degradation.
    \item \textbf{Delamination:} insertion of a thin ($\sim$18\,mm) air gap layer within the concrete matrix, representing bond failure between rebar and surrounding material.
\end{itemize}

A sixth class, healthy concrete with an intact cage, provides negative examples. In total, 900 volumes were generated: 150 per class (five defect types plus healthy baseline). Each volume contains 5,000 muon events, yielding 4.5 million events across the full dataset.

\subsection{Cloud Orchestration}

Simulation jobs were containerised and submitted to AWS Batch. Each job ran independently on a single vCPU, writing hit-level and summary-level CSV files to Amazon S3. The 900-job campaign completed in approximately 3 hours of wall-clock time at a cost under \$30 USD. A separate validation campaign of 60 volumes (10 per class) was generated independently after model training to test generalisation.

\subsection{Voxelisation and Feature Extraction}

The $1\,\text{m}^3$ target volume is discretised into a $20 \times 20 \times 20$ grid of 50\,mm cubic voxels. For each voxel, two feature tensors are accumulated over all 5,000 events:

\textbf{Stream~1 (scattering kinematics, 9 channels):} projected scattering angles $\theta_x$, $\theta_y$, total deflection $\theta_{\text{total}}$, spatial displacements $\Delta x$, $\Delta y$, energy loss $\Delta E$, track length ratio, primary energy deposit, and an event-count channel.

\textbf{Stream~2 (shower multiplicity, 40 channels):} for each of the six detector planes, six features are extracted from secondary particles (TrackID $\neq 1$): electron count, gamma count, positron count, shower energy deposit, spatial spread ($\sigma_{xy}$), and time spread. Three aggregate features---shower asymmetry, energy deposit ratio, and total secondary count---plus a hit-count channel complete the tensor. Dead channels (secondary neutrons and protons, identically zero at 4\,GeV) were excluded after diagnostic analysis.

Ground-truth labels assign each voxel one of six integer classes based on geometric intersection with the Geant4 solid definitions: 0 (concrete), 1 (honeycombing), 2 (shear), 3 (corrosion), 4 (delamination), 5 (rebar).

\section{Method}
\label{sec:method}

\subsection{Architecture Overview}

\begin{figure}[t]
    \centering
    \includegraphics[width=\columnwidth]{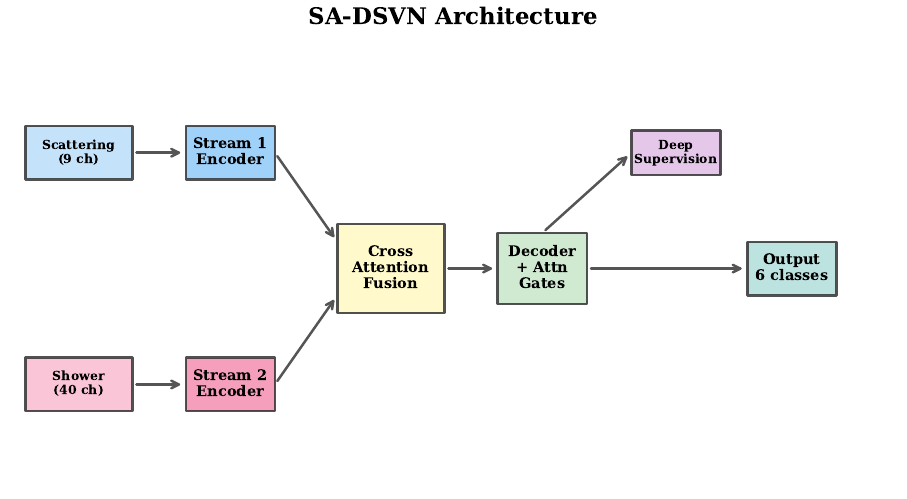}
    \caption{SA-DSVN architecture. The scattering stream (9 channels) and shower stream (40 channels) pass through independent 3-stage encoders. At the $5^3$ bottleneck, cross-attention fuses both representations. The decoder upsamples through attention-gated skip connections, with deep supervision at intermediate stages.}
    \label{fig:architecture}
\end{figure}

The SA-DSVN follows an encoder--decoder topology with three distinguishing features: dual-stream input processing, cross-attention fusion at the bottleneck, and attention-gated skip connections. The full model contains 1.87M trainable parameters. Fig.~\ref{fig:architecture} shows the dataflow.

\subsection{Dual-Stream Encoder}

Each input stream passes through an independent encoder consisting of three stages. Each stage applies a $3\times3\times3$ convolution, batch normalisation, ReLU activation, and $2\times$ max-pooling, progressively reducing spatial resolution from $20^3$ to $10^3$ to $5^3$. Stream~1 (9 input channels) processes scattering features. Stream~2 (40 input channels) processes shower features. The two streams share no weights.

The rationale for separation is physical. Scattering angles and shower multiplicities encode different aspects of the same interaction: the former reflects integrated Coulomb deflection along the muon path, the latter reflects local electromagnetic cascade intensity at discrete detector planes. Forcing the network to build independent representations before fusion prevents one modality from dominating early feature maps.

\subsection{Cross-Attention Fusion}

At the bottleneck ($5^3$ spatial resolution), the two encoded representations are fused via multi-head cross-attention with 4 heads. Stream~1 features serve as queries; Stream~2 features serve as keys and values. The attention output is added back to Stream~1 as a residual, then concatenated with Stream~2 before entering the decoder. Layer normalisation is applied before each projection.

This mechanism allows the scattering stream to selectively attend to shower features---for instance, querying whether a region of high scattering coincides with high secondary multiplicity (rebar) or low multiplicity (void). The asymmetry is deliberate: scattering provides the spatial ``where,'' and the shower stream disambiguates the ``what.''

\subsection{Decoder with Attention Gates}

The decoder mirrors the encoder with three upsampling stages using trilinear interpolation followed by $3\times3\times3$ convolution. Skip connections from the Stream~1 encoder are gated by attention gates~\cite{oktay2018}: each skip feature map is element-wise multiplied by a learned sigmoid mask conditioned on the decoder state. This suppresses encoder features in regions where the decoder has already resolved the class, reducing false activations at material boundaries.

A final $1\times1\times1$ convolution maps the decoded features to 6 output channels (one per class), producing a $20^3 \times 6$ logit volume.

\subsection{Deep Supervision}

Auxiliary classification heads are attached at each decoder stage, producing intermediate predictions at $5^3$ and $10^3$ resolution. These are upsampled to $20^3$ and compared against the ground truth during training. The auxiliary losses (weighted at 0.3 and 0.15) act as gradient highways to the encoder, stabilising training when the primary loss plateaus. At inference time, only the final output head is used.

\subsection{Loss Function}

The training objective combines focal loss~\cite{lin2017focal} and Dice loss:
\begin{equation}
    \mathcal{L} = \mathcal{L}_{\text{focal}}(\gamma=3) + \lambda_{\text{dice}} \cdot \mathcal{L}_{\text{dice}}
\end{equation}
where $\lambda_{\text{dice}} = 2.0$. Focal loss with $\gamma = 3$ strongly down-weights well-classified concrete voxels (which constitute $>$90\% of the volume), directing gradient signal toward rare defect boundaries. Per-class weights further compensate for imbalance: $w = [0.1, 20, 25, 20, 25, 0.5]$ for concrete, honeycombing, shear, corrosion, delamination, and rebar respectively.

\subsection{Training Protocol}

The 900 volumes are split 70/15/15 into train (628), validation (134), and test (138) sets with stratified sampling. All models are trained for 100 epochs with AdamW ($\beta_1 = 0.9$, $\beta_2 = 0.999$, weight decay $= 10^{-4}$) and batch size 4. The learning rate follows a linear warmup over 5 epochs to $10^{-3}$, then cosine annealing to $10^{-5}$. Early stopping with patience 30 is applied on validation Dice. Gradient norms are clipped at 1.0.

\subsection{Data Augmentation}

Three spatial augmentations are applied stochastically during training, each with probability 0.5: independent flips along the $x$, $y$, and $z$ axes. These multiply the effective training set by up to $8\times$ without altering the physics (the voxel grid has cubic symmetry). Intensity perturbations---Gaussian noise ($\sigma = 0.05$) and per-channel scaling ($\times\,[0.95, 1.05]$)---simulate detector-level variation. Augmentation is applied only to the training split; validation and test data are processed without modification.

As shown in Section~\ref{sec:aug}, augmentation is not optional. Without it, the model overfits to the training distribution and fails catastrophically on independently generated data.

\section{Experiments}
\label{sec:experiments}

\subsection{Ablation Study}
\label{sec:ablation}

\begin{figure}[!t]
    \centering
    \includegraphics[width=\columnwidth]{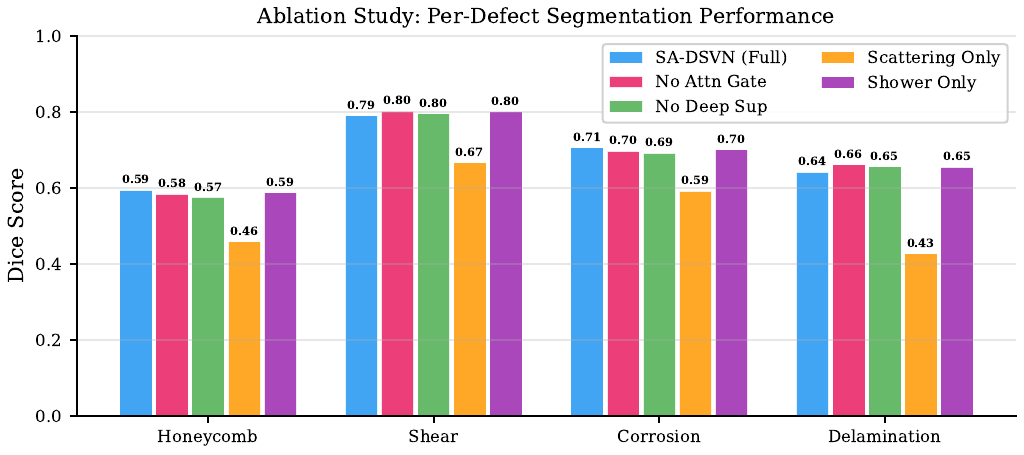}
    \caption{Per-defect Dice scores across the five ablation variants. The scattering-only model falls substantially behind all configurations that include shower data.}
    \label{fig:ablation_bars}
\end{figure}

To isolate the contribution of each architectural component, five model variants were trained under identical conditions (same data splits, augmentation, loss function, and hyperparameters). Table~\ref{tab:ablation} reports validation-set Dice scores and Fig.~\ref{fig:ablation_bars} visualises the per-defect breakdown.

\begin{table}[!t]
\centering
\caption{Ablation study. All models trained for 100 epochs with augmentation. Best values per column in \textbf{bold}.}
\label{tab:ablation}
\small
\setlength{\tabcolsep}{3pt}
\begin{tabular}{@{}lcccccc@{}}
\toprule
\textbf{Variant} & \textbf{Params} & \textbf{Overall} & \textbf{Honey.} & \textbf{Shear} & \textbf{Corr.} & \textbf{Delam.} \\
\midrule
SA-DSVN (Full)       & 1.87M & 0.945 & \textbf{0.594} & 0.790 & \textbf{0.705} & 0.641 \\
No Attn Gate         & 1.85M & \textbf{0.945} & 0.583 & \textbf{0.801} & 0.696 & \textbf{0.661} \\
No Deep Sup.         & 1.87M & 0.944 & 0.575 & 0.796 & 0.691 & 0.655 \\
Scat.\ Only          & 1.26M & 0.920 & 0.458 & 0.666 & 0.591 & 0.426 \\
Shower Only          & 1.27M & 0.945 & 0.586 & 0.801 & 0.701 & 0.654 \\
\bottomrule
\end{tabular}
\end{table}

\begin{figure}[!t]
    \centering
    \includegraphics[width=\columnwidth]{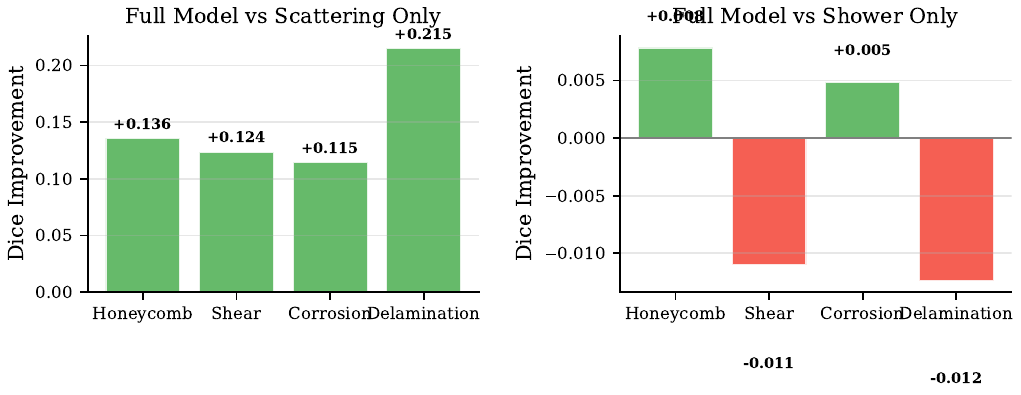}
    \caption{Per-defect Dice improvement when adding each stream to the baseline. The shower stream provides the bulk of discriminative power; scattering adds a smaller but consistent contribution, especially for honeycombing.}
    \label{fig:stream_delta}
\end{figure}

Three findings emerge:

\textit{The shower stream carries the dominant signal.} The Scattering Only variant drops defect-mean Dice by 28\% relative to the full model (0.535 vs.\ 0.683). Shower Only matches the full model within 0.2\%. This confirms that secondary multiplicity, not scattering angle, is the primary discriminant for defect detection in the presence of rebar. Fig.~\ref{fig:stream_delta} quantifies the per-class contribution of each stream.

\textit{Attention gates and deep supervision provide marginal gains.} Removing either component changes overall Dice by less than 0.2\%. The architecture is robust to these ablations, suggesting that the representational capacity of the dual-stream encoder is sufficient without explicit gating.

\textit{Scattering still helps for distributed defects.} The full model outperforms Shower Only on honeycombing (0.594 vs.\ 0.586), the most spatially distributed defect class. Scattering angles provide complementary long-range path information that shower multiplicity---measured at discrete detector planes---cannot capture alone.

\subsection{Training Dynamics}

\begin{figure}[!t]
    \centering
    \includegraphics[width=\columnwidth]{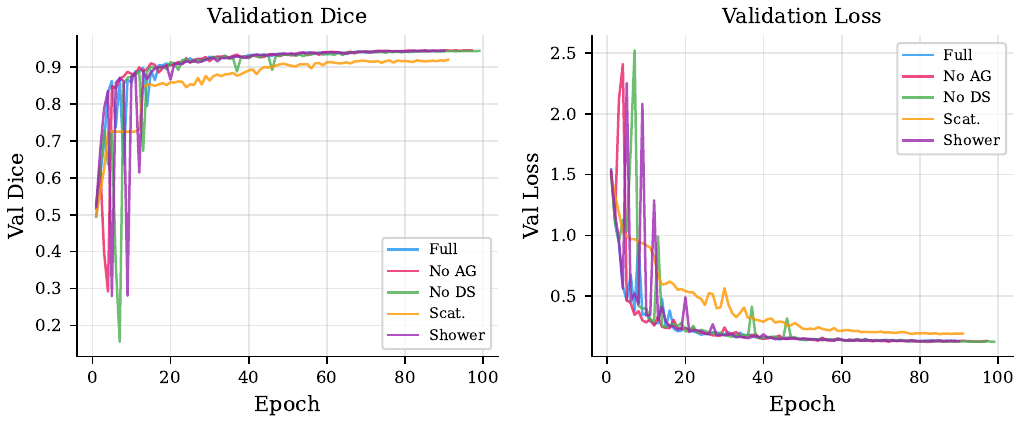}
    \caption{Validation Dice and training loss curves for all five ablation variants over 100 epochs. Dual-stream models converge to higher ceilings than the scattering-only variant.}
    \label{fig:training_curves}
\end{figure}

All five variants converge within 100 epochs (Fig.~\ref{fig:training_curves}). The dual-stream models reach a validation Dice plateau near epoch 70--80, while the scattering-only model converges faster but to a lower ceiling. The small train--validation gap ($<$0.05 Dice for all dual-stream models) indicates that overfitting is controlled by the augmentation regime.

\subsection{Validation-Set Performance}

The best model (SA-DSVN Full, selected by validation Dice) achieves the per-class scores shown in Table~\ref{tab:valset}. Concrete and rebar---the two high-volume classes---are segmented near-perfectly (Dice $>$ 0.97). Defect classes range from 0.58 (honeycombing) to 0.79 (shear). The lower honeycombing score reflects the difficulty of the class: small, randomly scattered voids that occupy fewer than 1\% of the volume.

\begin{table}[!t]
\centering
\caption{Per-class Dice and IoU on the validation set (134 volumes, same distribution as training).}
\label{tab:valset}
\small
\begin{tabular}{@{}lcc@{}}
\toprule
\textbf{Class} & \textbf{Dice} & \textbf{IoU} \\
\midrule
Concrete      & 0.985 & 0.972 \\
Honeycombing  & 0.594 & 0.423 \\
Shear         & 0.790 & 0.653 \\
Corrosion     & 0.705 & 0.545 \\
Delamination  & 0.641 & 0.472 \\
Rebar         & 0.977 & 0.958 \\
\midrule
\textit{Overall} & \textit{0.945} & --- \\
\textit{Defect mean} & \textit{0.683} & \textit{0.523} \\
\bottomrule
\end{tabular}
\end{table}

\subsection{Generalisation to Unseen Data}
\label{sec:fresh}

The decisive test of any learned model is performance on data from outside the training distribution. We generated 60 new volumes (10 per class) using an independent Vega simulation campaign with fresh random seeds, processed them through the same feature extraction pipeline with training-set normalisation statistics, and evaluated without any fine-tuning.

Table~\ref{tab:freshval} reports the results. Two outcomes stand out. First, voxel-level Dice scores are consistent with---and in some cases exceed---the validation-set numbers (e.g., shear 0.807 fresh vs.\ 0.790 validation). The model has not memorised the training volumes. Second, volume-level detection sensitivity is 100\% for all four defect classes: every defective volume is correctly identified as containing a defect, with AUC = 1.0 across the board.

\begin{table}[!t]
\centering
\caption{Fresh validation on 60 independently generated volumes (10 per class). No fine-tuning or adaptation applied. Volume-level metrics are one-vs-rest binary detection.}
\label{tab:freshval}
\small
\setlength{\tabcolsep}{3pt}
\begin{tabular}{@{}lccccc@{}}
\toprule
\textbf{Class} & \textbf{Dice} & \textbf{Sens.} & \textbf{Spec.} & \textbf{Prec.} & \textbf{AUC} \\
\midrule
Concrete      & 0.972 & ---   & ---  & ---  & ---  \\
Honeycombing  & 0.588 & 1.00 & 0.80 & 0.50 & 1.00 \\
Shear         & 0.807 & 1.00 & 0.90 & 0.67 & 1.00 \\
Corrosion     & 0.698 & 1.00 & 1.00 & 1.00 & 1.00 \\
Delamination  & 0.681 & 1.00 & 1.00 & 1.00 & 1.00 \\
Rebar         & 0.950 & ---   & ---  & ---  & ---  \\
\midrule
\textit{Overall accuracy} & \multicolumn{5}{c}{\textit{96.3\%}} \\
\bottomrule
\end{tabular}
\end{table}

\begin{figure}[!t]
    \centering
    \includegraphics[width=0.85\columnwidth]{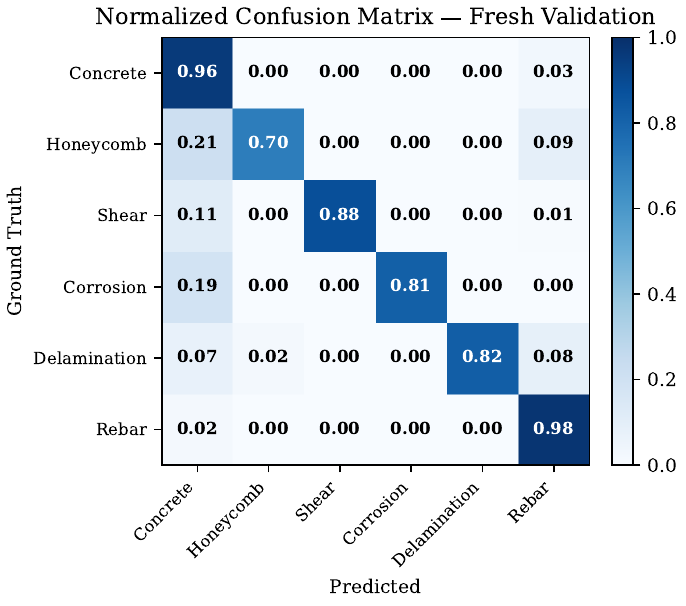}
    \caption{Normalised confusion matrix on the 60-volume fresh validation set. Off-diagonal mass concentrates at defect--concrete boundaries, indicating correct localisation with boundary uncertainty.}
    \label{fig:confusion_matrix}
\end{figure}

The confusion matrix (Fig.~\ref{fig:confusion_matrix}) reveals that the primary error mode is misclassification between defect and concrete at boundary voxels---the model correctly localises defect regions but produces slightly ``fuzzy'' borders. This is expected at 50\,mm voxel resolution, where partial-volume effects are significant.

\begin{figure}[!t]
    \centering
    \includegraphics[width=0.85\columnwidth]{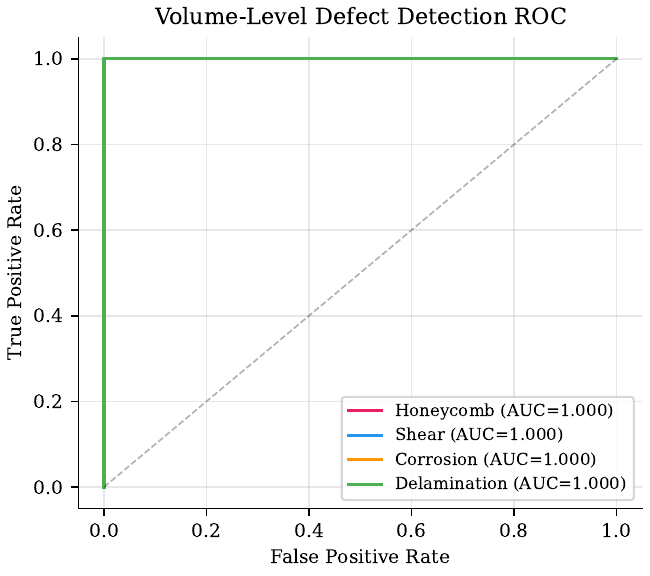}
    \caption{Volume-level ROC curves for defect detection on the fresh validation set. All four classes achieve AUC = 1.000.}
    \label{fig:roc_curves}
\end{figure}

Volume-level ROC curves (Fig.~\ref{fig:roc_curves}) show AUC = 1.000 for all four defect classes, meaning the model's continuous defect-probability scores rank every positive volume above every negative one. However, this does not imply error-free classification at a fixed threshold: honeycombing incurs 10 false positives out of 50 negatives (specificity 0.80), and shear incurs 5 (specificity 0.90). The model over-predicts defect presence in some healthy volumes but assigns them lower confidence than genuinely defective ones. Additionally, the small evaluation set (10 positives per class) means these AUC estimates carry wide confidence intervals; validation on a larger corpus is needed before drawing strong conclusions about volume-level reliability.

\begin{figure*}[!t]
    \centering
    \includegraphics[width=0.85\textwidth]{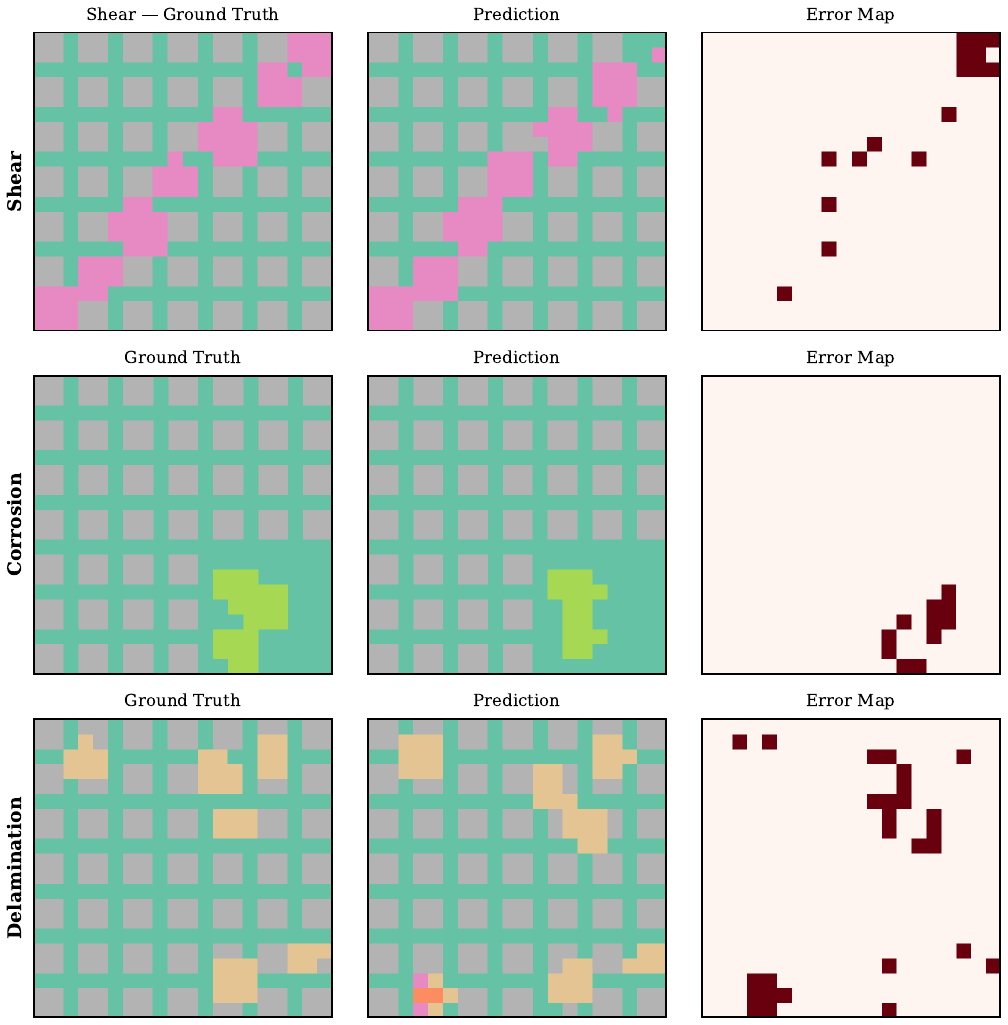}
    \caption{Representative 2D slices through three validation volumes. Left column: ground truth. Centre: model prediction. Right: error map (coloured pixels indicate misclassified voxels). Boundary errors are visible but defect regions are spatially correct.}
    \label{fig:slice_comparison}
\end{figure*}

Qualitative slice comparisons (Fig.~\ref{fig:slice_comparison}) show that the predicted segmentation closely matches ground truth across representative volumes. Defect regions are correctly localised, rebar bars are preserved, and errors are confined to 1--2 voxel-wide boundary zones.

\subsection{The Role of Augmentation}
\label{sec:aug}

\begin{figure}[!t]
    \centering
    \includegraphics[width=\columnwidth]{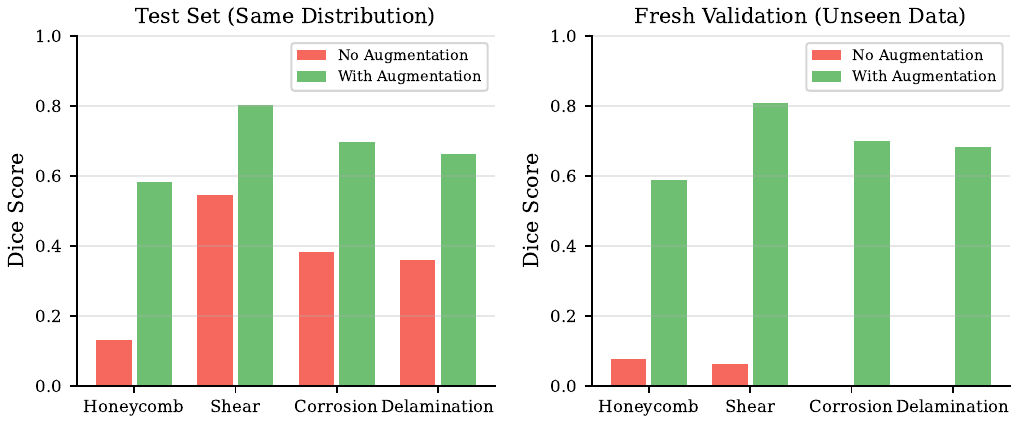}
    \caption{Effect of augmentation on generalisation. Test-set performance (same distribution as training) is comparable, but fresh-data performance collapses without augmentation---demonstrating that augmentation is essential for voxelised muon tomography.}
    \label{fig:augmentation_impact}
\end{figure}

To quantify the impact of data augmentation, we compare two models trained on identical data: one with augmentation (3D flips + intensity noise), one without. Table~\ref{tab:aug} reports Dice on the 60 fresh validation volumes, and Fig.~\ref{fig:augmentation_impact} illustrates the contrast graphically.

Without augmentation, the model achieves strong Dice on the training-distribution test set (shear 0.55, corrosion 0.38) but collapses entirely on fresh data. Corrosion and delamination produce zero Dice---the model predicts only concrete for these volumes. With augmentation, all four classes recover to Dice $\geq$ 0.58 on fresh data. The $8\times$ expansion of the effective training set through axis flips breaks the spatial priors that the unaugmented model memorises, forcing it to learn orientation-invariant features.

\begin{table}[!t]
\centering
\caption{Effect of augmentation on generalisation. Dice scores on 60 fresh validation volumes.}
\label{tab:aug}
\small
\begin{tabular}{@{}lcc@{}}
\toprule
\textbf{Defect} & \textbf{No Aug} & \textbf{With Aug} \\
\midrule
Honeycombing  & 0.076 & 0.588 \\
Shear         & 0.063 & 0.807 \\
Corrosion     & 0.000 & 0.698 \\
Delamination  & 0.000 & 0.681 \\
\bottomrule
\end{tabular}
\end{table}

\subsection{Computational Cost}

Inference on a single $20^3$ volume takes $10.4 \pm 2.8$\,ms on an Apple M-series GPU (Metal Performance Shaders). Training the full model for 100 epochs requires approximately 90 minutes on the same hardware. The simulation campaign of 900 volumes cost under \$30 USD on AWS Batch. The complete pipeline---simulation, voxelisation, training, and inference---can be reproduced for under \$50 in cloud compute.

\subsection{Failure Modes}

Honeycombing remains the weakest class (Dice 0.588). The defect consists of many small, scattered voids that individually span only 1--2 voxels at 50\,mm resolution. The model correctly localises the affected region but over-predicts its extent, producing false positives in adjacent concrete voxels. Higher grid resolution ($40^3$ or $64^3$) would likely improve boundary precision but at $8\times$ to $32\times$ the memory cost.

Delamination presents a different challenge. At 18\,mm physical thickness, a delamination layer is thinner than the 50\,mm voxel size. The model detects its presence (100\% sensitivity) and approximate location (centroid error $<$ 1 voxel) but cannot resolve its exact thickness. This is a resolution limit, not a model failure.

\section{Conclusion}
\label{sec:conclusion}

Secondary electromagnetic shower multiplicity is an effective learned feature for muon tomographic reconstruction. A dual-stream network that processes shower data alongside scattering kinematics segments four types of structural defects in reinforced concrete with Dice scores of 0.59--0.81 on independently generated validation data, while detecting defect presence with 100\% sensitivity. The shower stream alone accounts for most of this performance; scattering provides a complementary but secondary contribution.

Data augmentation proved essential. Without it, the same architecture generalises poorly despite strong in-distribution performance---a cautionary result for future simulation-trained models in this domain.

Several limitations remain. The 50\,mm voxel resolution cannot resolve thin features such as delamination layers or micro-voids smaller than the voxel pitch. All experiments use simulated data with mono-energetic muons at normal incidence; real cosmic-ray spectra are broad and anisotropic. Validation on physical detector data is the necessary next step.

Future work will address three directions: (1) increasing grid resolution to $40^3$ or $64^3$ with memory-efficient architectures, (2) replacing the mono-energetic beam with a realistic cosmic-ray energy spectrum and angular distribution, and (3) deploying the trained model against empirical data from a physical muon tomography detector to validate simulation-to-reality transfer. The Vega simulation framework and trained model weights will be made publicly available to support reproducibility.

\section*{Acknowledgement}

The authors thank the Geant4 Collaboration for the open-source physics simulation libraries that underpin the Vega framework. Cloud computing resources were provided by Amazon Web Services.

\bibliographystyle{IEEEtran}
\bibliography{references}

\end{document}